\documentclass{article}
\usepackage{microtype}
\usepackage{graphicx}
\usepackage{subfigure}
\usepackage{booktabs}
\usepackage{amsfonts}
\usepackage{amssymb}  
\usepackage{amsthm}
\usepackage{amsmath}
\usepackage{enumitem}
\usepackage{dsfont}

\newcommand{\defeq}{:=}
\newcommand{\norm}[1]{\left\lVert#1\right\rVert}

\usepackage{hyperref}
\usepackage{bbm}

\usepackage[accepted]{icml2019}

\icmltitlerunning{Learning Generative Models across Incomparable Spaces}

\begin{document}

\twocolumn[
\icmltitle{Learning Generative Models across Incomparable Spaces}
\icmlsetsymbol{equal}{*}

\begin{icmlauthorlist}
\icmlauthor{Charlotte Bunne}{eth}
\icmlauthor{David Alvarez-Melis}{mit}
\icmlauthor{Andreas Krause}{eth}
\icmlauthor{Stefanie Jegelka}{mit}
\end{icmlauthorlist}

\icmlaffiliation{eth}{Department of Computer Science, Eidgen\"{o}ssische Technische Hochschule (ETH), Z\"{u}rich, Switzerland}
\icmlaffiliation{mit}{Computer Science and Artificial Intelligence Laboratory (CSAIL), Massachusetts Institute of Technology (MIT), Cambridge, USA}

\icmlcorrespondingauthor{Charlotte Bunne}{bunnec@ethz.ch}

\icmlkeywords{Generative Models, Gromov-Wasserstein Divergence, Orthogonal Procrustes-based Regularization}
\vskip 0.3in
]

\printAffiliationsAndNotice{}

\begin{abstract}

Generative Adversarial Networks have shown remarkable success in learning a distribution that faithfully recovers a reference distribution \textit{in its entirety}. However, in some cases, we may want to only learn some aspects (e.g., cluster or manifold structure), while modifying others (e.g., style, orientation or dimension). In this work, we propose an approach to learn generative models across such \textit{incomparable} spaces, and demonstrate how to steer the learned distribution towards target properties. A key component of our model is the \emph{Gromov-Wasserstein} distance, a notion of discrepancy that compares distributions \textit{relationally} rather than absolutely. While this framework subsumes current generative models in identically reproducing distributions, its inherent flexibility allows application to tasks in manifold learning, relational learning and cross-domain learning.

\end{abstract}

\section{Introduction}
Generative Adversarial Networks (GANs, \citet{Goodfellow2014}) and its variations \citep{Radford2015, Arjovsky2017, Li2017} are powerful models for learning complex distributions. Broadly, these methods rely on an \textit{adversary} that compares 
samples from the true and learned distributions, giving rise to a notion of divergence between them. The divergences implied by current methods require the two distributions to be supported in sets that are \textit{identical} or at the very least \textit{comparable}; examples include 
Optimal Transport (OT) distances \citep{Salimans2018, Genevay2018} or Integral Probability Metrics (IPM) \citep{Muller1997, Sriperumbudur2012, Mroueh2017}. In all of these cases, the spaces over which the distributions are defined must have the same dimensionality (e.g., the space of $28\times 28$-pixel vectors for MNIST), and the generated distribution that minimizes the objective has the same support as the reference one. This is of course desirable when the goal is to generate samples that are \textit{indistinguishable} from those of the reference distribution. 

Many other applications, however, require modeling only topological or relational aspects of the reference distribution. In such cases, the absolute location of the data manifold is irrelevant (e.g., distributions over learned representations, such as word embeddings, are defined only up to rotations), or it is not available (e.g., if the data is accessible only as a weighted graph indicating similarities among sample points). Another reason for modeling only topological aspects is the desire to, e.g., change the appearance or style of the samples, or down-scale images.
Divergences that directly compare samples from the two distributions, and hence most current generative models, do not apply to those settings.

\begin{figure*}[t]
  \includegraphics[width=.99\textwidth]{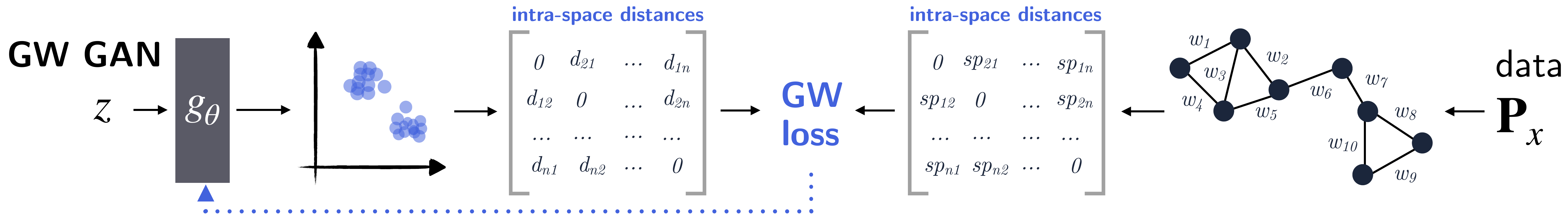}
  \caption{The Gromov-Wasserstein generative adversarial network (\textsc{Gw Gan}) learns across incomparable spaces, such as different dimensions or data type (from graphs to Euclidean space). The key idea is that its learning objective is purely based on intra-space distances (e.g., pairwise distances $d$ or shortest paths $sp$) in the generator and data space, respectively.}
  \label{fig:principle}
\end{figure*}

In this work, we develop a novel class of generative models that can learn across \textit{incomparable} spaces, e.g., spaces of different dimensionality or data type. Here, the relational information between samples, i.e., the topology of the reference data manifold, is preserved, but other 
characteristics, such as the ambient dimension, can vary.
A key component of our approach is the \emph{Gromov-Wasserstein} ($\text{GW}$) distance \citep{Memoli2011}, a generalization of classic Optimal Transport distances to incomparable ground spaces. Instead of directly comparing points in the two spaces, the $\text{GW}$ distance computes pairwise intra-space distances, and compares those distances across spaces, greatly increasing the modeling scope. Figure~\ref{fig:principle} illustrates the new model.

To realize this model, we address several challenges. First, we enable the use of the Gromov-Wasserstein distance in various learning settings by improving its robustness and ensuring unbiased learning. Similar to existing OT-based generative models \citep{Salimans2018, Genevay2018}, we leverage the differentiability of this distance to provide gradients for the generator. Second, for efficiency, we further parametrize it via a learnable adversary. The added flexibility of the $\text{GW}$ distance necessitates to constrain the adversary. To this end, we propose a novel 
orthogonality regularization, which might be of independent interest.

A final challenge ---which doubles as one of the main advantages of this approach--- arises from the added flexibility of the generator: it allows to freely alter superficial characteristics of the generated distribution while still learning the basic structure of the reference distribution. We show examples how to steer these additional degrees of freedom via regularization or adversaries in the model.
The resulting model subsumes the traditional (i.e., same-space) adversarial models as a special case, but can do much more. For example, it learns cluster structure across spaces of different dimensionality and across different data types, e.g., from graphs to Euclidean space. Thus, our \textsc{Gw Gan} can also be viewed as performing dimensionality reduction or manifold learning, but, departing from classical approaches to these problems, it recovers, in addition to the manifold structure of the data, the probability distribution defined over it.
Moreover, we propose a general framework for stylistic modifications by integrating a style adversary; we demonstrate its use by changing the thickness of learned MNIST digits.
In summary, this work provides a framework that substantially expands the potential applications of generative adversarial learning.

\vspace{-7pt}
\paragraph{Contributions.} We make the following contributions:
\begin{enumerate}[noitemsep, nolistsep, label=\roman*.]
  \vspace{-5pt}
	\item We introduce a new class of generative models that can learn distributions across different dimensionalities or data types.
	\item We demonstrate the model's range of applications by deploying it to manifold learning, relational learning and cross-domain learning tasks.
	\item More generally, our modifications of the Gromov-Wasserstein discrepancy enable its use as a loss function in various machine learning applications.
	\item Our new approach to approximately enforce orthogonality in neural networks based on the orthogonal Procrustes problem also applies beyond our model.
\end{enumerate}


\section{Model}
Given a dataset of $n$ observations $\{ x_1, ..., x_n\}, x_i \in \mathcal{X}$ drawn from a reference distribution $p \in \mathcal{P}(\mathcal{X})$, we aim to learn a generative model $g_\theta$ parametrized by $\theta$ 
purely based on relational and intra-structural characteristics of the dataset. The generative model $g_\theta: \mathcal{Z} \rightarrow \mathcal{Y}$, typically a neural network, maps random noise $z \in \mathcal{Z}$ to a generator space $\mathcal{Y}$ that is independent of data space $\mathcal{X}$. 

\begin{algorithm*}[t]
\caption{Training Algorithm of the Gromov-Wasserstein Generative Model.}\label{alg:gwgan}
\begin{algorithmic}
\REQUIRE  $\alpha$: learning rate, $n_g$: the number of iterations of the generator per adversary iteration, $m$: mini-batch size, \\ $N$: number of training iterations, $\theta_0$: initial parameters of generator $g_\theta$, $\omega_0 = (\check{\omega}_0, \hat{\omega}_0)$: initial parameters of adversary $f_\omega$
\FOR{$t = 0$ to $N$}
	\STATE sample $X = (x_i)^m_{i=1}$ from dataset
	\STATE sample $Z = (z_j)^m_{j=1} \sim \mathcal{N}(0,1)$, $Y = (y_j)^m_{j=1} = g_{\theta_t}((z_j)^m_{j=1})$
	\STATE $\forall (i, j), D^{\check{\omega}}_{ij} \defeq \norm{f_{\check{\omega}}(x_i) - f_{\check{\omega}}(x_j)}_2$ and $D^{\hat{\omega}}_{ij} \defeq \norm{f_{\hat{\omega}}(y_i) - f_{\hat{\omega}}(y_j)}_2$
	\STATE $\mathcal{L} = \overline{\text{GW}}_\epsilon(D^{\check{\omega}}, D^{\hat{\omega}}, \mathbf{p}, \mathbf{q})$, where $\mathbf{p}, \mathbf{q}$ are uniform distributions \COMMENT{$\overline{\text{GW}}_\epsilon$ is defined in Eq.~\eqref{eq:gw_norm}}
	\IF{$t \mod n_g + 1 = 0$}
		\STATE $\mathcal{L}_{\mathrm{reg}} \leftarrow \mathcal{L} - R_\beta(f_{\check{\omega}}(X), X) - R_\beta(f_{\hat{\omega}}(Y), Y)$ \COMMENT{$R_\beta$ is defined in Eq.~\eqref{eq:reg_procrustes}}
		\STATE $\omega_{t+1} \leftarrow  \omega_{t} + \alpha \times \nabla_{\omega_{t}} \mathcal{L}_{\mathrm{reg}}$
	\ELSE
		\STATE $\theta_{t+1} \leftarrow  \theta_{t} - \alpha \times \nabla_{\theta_{t}} \mathcal{L}$
	\ENDIF
\ENDFOR
\end{algorithmic}
\end{algorithm*}

\subsection{Gromov-Wasserstein Discrepancy}
Learning generative models typically relies 
on a statistical divergence between the target distribution and the model's current estimate.
Classical statistical divergences only apply when comparing distributions whose supports lie in the same metric space, or when at least a meaningful distance between points in the two supports can be computed. When the data space $\mathcal{X}$ and generator space $\mathcal{Y}$ are different, these divergences no longer apply.

Hence, instead, we will use a more suitable divergence measure. 
Rather than relying on a metric \textit{across} the spaces, the \emph{Gromov-Wasserstein ($\text{GW}$)} distance \citep{Memoli2011} compares distributions by computing a discrepancy between the metrics defined \textit{within} each of the spaces. As a consequence, it is oblivious to specific characteristics or the dimensionality of the spaces.

Given $n$ samples of the compared distributions $p \in \mathcal{P}(\mathcal{X})$ and $q \in \mathcal{P}(\mathcal{Y})$, 
the discrete formulation of the $\text{GW}$ distance needs 
a similarity (or distance) matrix between the samples and a probability vector 
for each space, say $(D, \mathbf{p})$ and $(\bar{D}, \mathbf{q})$, with $(D, \mathbf{p}) \in \mathbb{R}^{n \times n} \times \Sigma _n$, where $\Sigma _n \defeq \{\mathbf{p} \in \mathbb{R}^+_n; \sum_i \mathbf{p}_i = 1\}$ is the $n$-dimensional probability simplex. 
Then, the $\text{GW}$ discrepancy is 
\begin{equation}\label{eq:gw}
\begin{split}
    \text{GW}(D, \bar{D}, \mathbf{p}, \mathbf{q}) &\defeq \min_{T \in \mathcal{U}_{\mathbf{p},\mathbf{q}}} \mathcal{E} _{D, \bar{D}} (T) \\
    &\defeq  \min_{T \in \mathcal{U}_{\mathbf{p},\mathbf{q}}} \sum_{ijkl} L( D_{ik}, \bar{D}_{jl}) T_{ij} T_{kl},
\end{split}
\end{equation}
where $\mathcal{U}_{\mathbf{p}, \mathbf{q}} \defeq \{ T \in (\mathbb{R}_+ )^{n \times n}; T \mathds{1}_n = \mathbf{p}, T^\top \mathds{1}_n = \mathbf{q}\}$ is the set of all couplings $T$ between $\mathbf{p}$ and $\mathbf{q}$. The loss function $L$ in our case is $L(a, b) = L_2(a, b) \defeq \frac{1}{2} |a - b|^2$.
If $L=L_2$, then $\text{GW}^{1/2}$ defines a (true) distance \citep{Memoli2011}. 


\subsection{Gromov-Wasserstein Generative Model}
To learn across incomparable spaces, one key idea of our model is to use the Gromov-Wasserstein distance as a loss function to compare the generated and true distribution.
As in traditional adversarial approaches, we parametrize the generator $g_\theta: \mathcal{Z} \rightarrow \mathcal{Y}$ as a neural network that maps noise samples $z$ to features $y$. We train $g_\theta$ by using $\text{GW}$ as a loss, i.e., for mini-batches $X$ and $Y$ of reference and generated samples, respectively, we compute pairwise distance matrices $D$ and $\bar{D}$ and solve the $\text{GW}$ problem, taking $\mathbf{p}$ and $\mathbf{q}$ as uniform distributions. 

While this procedure alone is often sufficient for simple problems, in high dimensions, the statistical efficiency of classical divergence measures 
can be poor and a large number of input samples is needed to achieve good discrimination between generated and data distribution \citep{Salimans2018}. To improve discriminability, we learn the intra-space metrics adversarially. An adversary $f_\omega$ parametrized by $\omega$ maps 
data and generator samples into feature spaces in which we compute Euclidean intra-space distances:
\begin{equation}\label{eq:map}
	D^\omega_{ij} \defeq \norm{f_\omega(x_i) - f_\omega(x_j)}_2, \textrm{ where } f_\omega \colon \mathcal{X} \rightarrow \mathbb{R}^s
\end{equation}
with $f_\omega$ modeled by a neural network. 
The feature mapping may, for instance, reduce the dimensionality of $\mathcal{X}$ and 
extract important features.
The original loss minimization problem of the generator 
thus becomes a minimax problem
\begin{equation}\label{eq:minmax}
	\min_\theta \max_{\omega = (\check{\omega}, \hat{\omega})} \quad \text{GW} (D^{\check{\omega}}, D^{\hat{\omega}}, \mathbf{p}, \mathbf{q}),
\end{equation}
where $D^{\check{\omega}}$ and $D^{\hat{\omega}}$ denote pairwise distance matrices of samples originating from the generator and reference domain, respectively, mapped into the feature space via $f_\omega$ (Eq.~\eqref{eq:map}). We refer to our model as \textsc{Gw Gan}.

\section{Training}
We optimize the adversary $f_\omega$ and generator $g_\theta$ in an alternating scheme, where we
 train the generator more frequently than the adversary to avoid the adversarially-learned distance function to become degenerate \citep{Salimans2018}. Algorithm~\ref{alg:gwgan} shows the \textsc{Gw Gan} training algorithm.

While training of standard GANs suffers from undamped oscillations and mode collapse \citep{Metz2017, Salimans2016}, following an argument of \citet{Salimans2018}, the $\text{GW}$ objective 
is well defined and statistically consistent if the adversarially learned intra-space distances $D^{\omega}$ are non-degenerate. Trained with the $\text{GW}$ loss, the generator thus does not diverge even when adversary $f_\omega$ is kept fixed.
Empirical validation (see Appendix~\ref{sec:fixed_ad}) confirms this: we stopped updating the adversary $f_\omega$ while continuing to update the generator. Even with fixed adversary, the generator further improved its learned distribution and did not diverge. 

Note that Problem \eqref{eq:minmax} makes very few assumptions on the spaces $\mathcal{X}$ and $\mathcal{Y}$, requiring only that a metric be defined on them. This remarkable flexibility can be exploited to enforce various characteristics on the generated distribution. We discuss examples in Section~\ref{sub:constr_gen}. However, this same flexibility combined with the added degrees of freedom due to the learned metric, demands to regularize the adversary to ensure stable training and prevent it from overpowering the generator. We propose an effective method to do so in Section~\ref{sub:reg_adv}.

Moreover, using the Gromov-Wasserstein distance as a differentiable loss function for training a generative model requires modifying its original formulation to ensure robust and fast computation, unbiased gradients, and numerical stability, as described in detail in Section~\ref{sec:gw_mod}.

\subsection{Constraining the Generator}\label{sub:constr_gen}
The $\text{GW}$ loss encourages the generator to recover the relational and geometric properties of the reference dataset, but leaves other global aspects undetermined. We can thus \textit{shape} the generated distribution by enforcing desired properties through constraints. For example, while any translation of a distribution would achieve the same $\text{GW}$ loss, we can enforce centering around the origin by penalizing the norm of the generated samples. Figure~\ref{fig:distributions}a illustrates an example.

For computer vision tasks, we need to ensure that the generated samples still look like natural images. We found that a total variation regularization \citep{Rudin1992} induces the right bias here and hence greatly improves the results (see Figures~\ref{fig:distributions}b, c, and d).

Moreover, the invariances of the $\text{GW}$ loss allow for shaping stylistic characteristics of the generated samples by integrating design constraints into the learning process.
In contrast, current generative models \citep{Arjovsky2017, Salimans2018, Genevay2018, Li2017} cannot perform style transfer as modifications in surface-level features of the generated samples conflict with their adversarially computed loss used for training the generator.
We propose a modular framework, which enables style transfer to the generated samples when given an additional style reference besides the provided data samples. We incorporate design constraints into the generator's objective via a \emph{style adversary} $c$, i.e., any function that quantifies a certain style and thereby, as a penalty, enforces this style on the generated samples. The resulting objective is
\begin{equation}\label{eq:loss_style}
	\min_\theta \max_{\omega = (\check{\omega}, \hat{\omega})} \quad \text{GW}(D^{\check{\omega}}, D^{\hat{\omega}}, \mathbf{p}, \mathbf{q}) - \lambda \times c(g_\theta(z)).
\end{equation}
As a result, the generator learns structural content of the target distribution via the adversarially learned $\text{GW}$ loss, and stylistic characteristics via the style adversary.
We demonstrate this framework via the example of stylistic changes to learned MNIST digits in Section~\ref{sec:style_bold} and in Appendix~\ref{sec:train_sa}.

\subsection{Regularizing the Adversary}\label{sub:reg_adv}
During training, the adversary maximizes the objective function \eqref{eq:minmax}.
However, the $\text{GW}$ distance is easily maximized by stretching the space and thus distorting the intra-space distances used for its computation.
To avoid such arbitrary distortion of the space, we propose to regularize the adversary $f_\omega$ by (approximately) enforcing it to define a unitary transformation, thus restricting the magnitude of stretching it can do. Note that \textit{directly} parametrizing $f_\omega$ as an orthogonal matrix would defeat its purpose, as the Frobenius norm is unitarily invariant. Instead, we allow $f_\omega$ to take a more general form, but limit its expansivity and contractivity through approximate orthogonality.

Previous work has explored various orthogonality-based regularization methods to 
stabilize neural networks training\citep{Vorontsov2017}. \citet{Saxe2013} introduced a new class of random orthogonal initial conditions on the weights of neural networks stabilizing the initial training phase. 
By enforcing the weight matrices to be Parseval tight frames, layerwise orthogonality constraints are introduced in \citet{Cisse2017, Brock2017, Brock2019}; they penalize deviations of the weights from orthogonality via $R_\beta(W_k) \defeq \beta \|W_k^\top W_k - I\|_F^2$, where $W_k$ are weights of layer $k$ and $\|\cdot\|_F$ is the Frobenius norm.

However, these approaches enforce orthogonality on the weights of each layer rather than constraining the network $f_\omega$ in its entirety to function as an approximately orthogonal operator. 
An empirical comparison to these layerwise approaches (shown in Appendix~\ref{sec:comp_orth}) reveals that, for \textsc{Gw Gan}, regularizing the full network is desirable.
To enforce the approximation of $f_\omega$ as an orthogonal operator, we introduce a new orthogonal regularization approach, which ensures orthogonality of a network by minimizing the distance to the closest orthogonal matrix $P^*$. The regularization term is defined as
\begin{equation}\label{eq:reg_procrustes}
	R_\beta(f_\omega(X), X) \defeq \beta \| f_\omega(X) - XP^{*\top} \|_F^2,
\end{equation}
where $P^*$ is an orthogonal matrix that most closely maps $X$ to  $f_\omega(X)$, and $\beta$ is a hyperparameter. The matrix $P^* = \arg \min_{P \in O(s)} \| f_\omega(X) - XP^\top \|_F$,
where $O(s) = \left \{ P \in \mathbb{R}^{s \times s} \; | \; P^\top P = I \right \}$ and $s$ is the dimensionality of the feature space, can be obtained by solving an orthogonal Procrustes problem. 
If the dimensionality $s$ of the feature space equals the input dimension, then $P^*$ has a closed-form solution $P^* = UV^\top$, where $U$ and $V$ are the left and right singular vectors of $f_\omega(X)^\top X$, i.e. $U \Sigma V^\top = \textrm{SVD}(f_\omega(X)^\top X)$ \citep{Schonemann1966}. Otherwise, we need to solve for $P^*$ with an iterative optimization method.
%

This novel Procrustes-based regularization principle for neural networks is remarkably flexible since it constrains global input-output behavior without making assumptions about specific layers or activations. 
It preserves the expressibility of the network while efficiently enforcing orthogonality. We use this orthogonal regularization principle for training the adversary of the $\text{GW}$ generative model across different applications and network architectures.

\subsection{Gromov-Wasserstein as a Loss Function}\label{sec:gw_mod}
To serve as a robust training objective  for general machine learning settings we modify the na\"ive formulation of the Gromov-Wasserstein discrepancy in various ways.

\paragraph{Regularization of Gromov-Wasserstein}
Optimal transport metrics and extensions such as the Gromov-Wasserstein distance are particularly appealing because they take into account the underlying geometry of the data when comparing distributions. However, their computational cost is prohibitive for large-scale machine learning problems. More precisely, Problem~\eqref{eq:gw} is a quadratic programming problem, and solving it directly is intractable for large $n$. 
Regularizing this objective with an entropy term results in significantly more efficient optimization  \citep{Peyre2016}. The resulting smoothed problem can be solved through projected gradient descent methods, where the projection steps rely on the Sinkhorn-Knopp scaling algorithm \citep{Cuturi2013}. Concretely, the entropy-regularized version of the Gromov-Wasserstein discrepancy proposed by \citet{Peyre2016} has the form
\begin{equation}\label{eq:gw_entropy}
    \text{GW}_{\epsilon}(D, \bar{D}, \mathbf{p}, \mathbf{q}) = \min_{T \in \mathcal{U}_{\mathbf{p},\mathbf{q}}} \mathcal{E} _{D, \bar{D}} (T) - \epsilon H(T),
\end{equation}
where $\mathcal{E} _{D, \bar{D}} (T)$ is defined in Equation~\eqref{eq:gw}, $H(T) \defeq - \sum_{ij} T_{ij} (\log (T_{ij}) -1)$ is the entropy of coupling $T$, and $\epsilon$ a parameter controlling the strength of regularization. Besides leading to significant speedups, entropy smoothing of optimal transport discrepancies results in distances that are \emph{differentiable} with respect to their inputs, making them a more convenient choice as loss functions for machine learning algorithms. 
Since the Gromov-Wasserstein distance as loss function in generative models compares noisy features of the generator and the data by computing correspondences between intra-space distances, a soft rather than a hard alignment $T$ might be a desirable property.
The entropy smoothing yields couplings that are sparser than their non-regularized counterparts, ideal for applications where soft alignments are desired \citep{Cuturi2016}. 

The effectiveness of entropy smoothing of the Gromov-Wasserstein discrepancy has been shown in other downstream application such as shape correspondences \citep{Solomon2016} or the alignment of word embedding spaces \citep{Melis2018}.

Motivated by \citet{Salimans2018} and justified by the envelope theorem \citep{Carter2001}, we do not backpropagate the gradient through the iterative computation of the $\text{GW}_\epsilon$ coupling $T$ (Problem \eqref{eq:gw_entropy}).

\paragraph{Normalization of Gromov-Wasserstein}
With entropy regularization, $\text{GW}_{\epsilon}$ is not a distance any more, as the discrepancy of identical metric measure spaces is then no longer zero. Similar to the Wasserstein metric \citep{Bellemare2017}, the estimation of $\text{GW}_\epsilon$ from samples yields biased gradients. Inspired by \citet{Bellemare2017}, we use a normalized entropy-regularized Gromov-Wasserstein discrepancy defined as
\begin{equation}\label{eq:gw_norm}
\begin{split}
    \overline{\text{GW}}_\epsilon(D, \bar{D}, \mathbf{p}, \mathbf{q}) \defeq  \ 2 \times \text{GW}_\epsilon(D, \bar{D}, \mathbf{p}, \mathbf{q})& 
    \\ - \text{GW}_\epsilon(D, D, \mathbf{p}, \mathbf{p}) - \text{GW}_\epsilon(\bar{D}, \bar{D}, \mathbf{q}, \mathbf{q})&.
\end{split}
\end{equation}

\begin{figure*}[t]
  \centering
  \includegraphics[width=.86\textwidth]{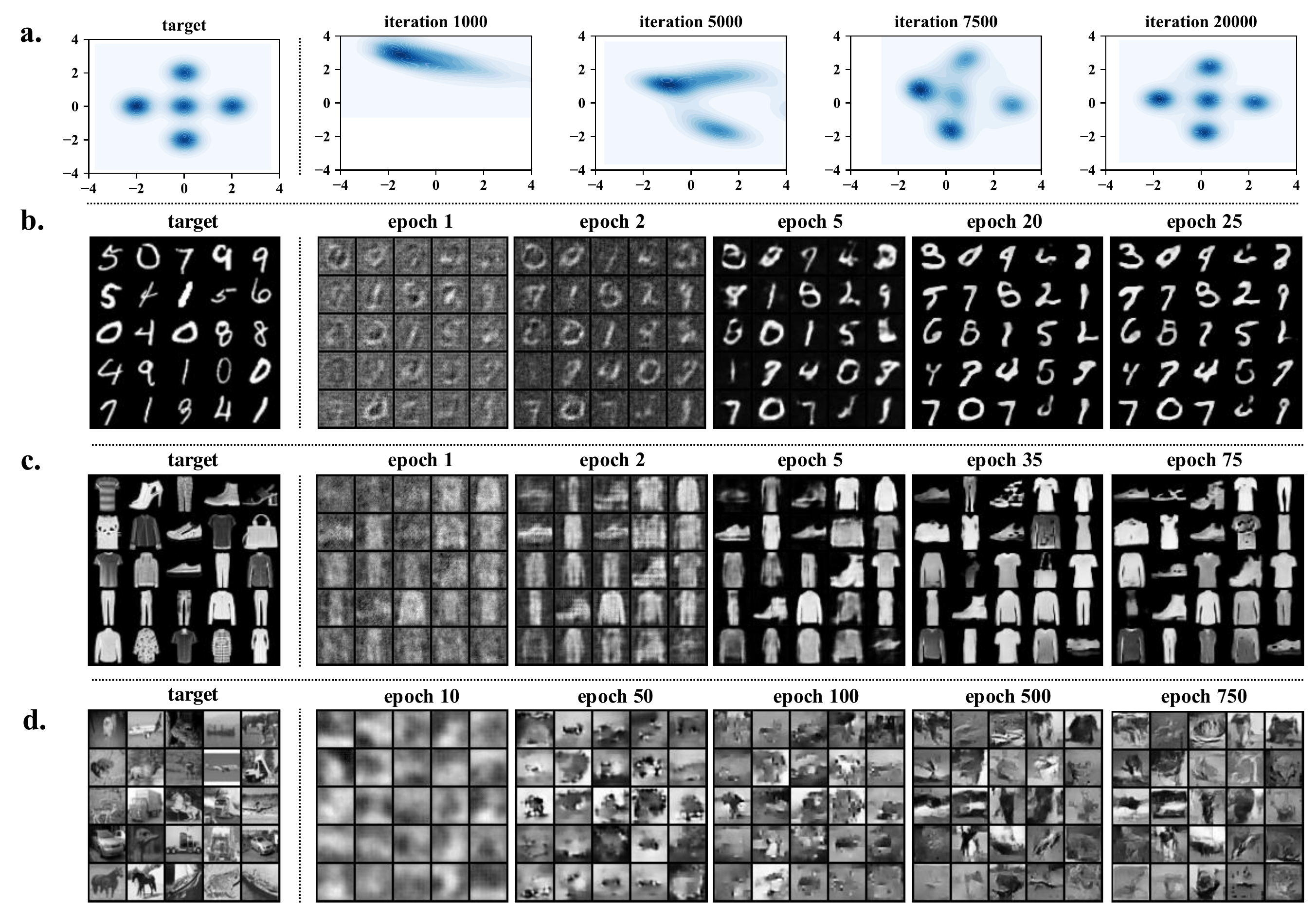}
  \caption{The Gromov-Wasserstein GAN can learn distributions of different dimensionality. \textbf{a.} Results of learning a mixture of Gaussian distributions with adversary $f_{\omega}$ ($\beta = 1$). $\ell_{1}$-regularization allows centering the distribution across the origin ($\ell_1$-penalty: $\lambda = 0.001$). Each plot shows 1000 generated samples. Learning to generate \textbf{b.} MNIST digits ($\beta = 32$), \textbf{c.} fashion-MNIST ($\beta = 35$) and \textbf{d.} gray-scale CIFAR10 ($\beta = 40$). Trained with total variation denoising ($\lambda = 0.5$).}
  \label{fig:distributions}
\end{figure*}

\paragraph{Numerical Stability of Gromov-Wasserstein}
Computing the entropy-regularized Gromov-Wasserstein formulation relies on a projected gradient algorithm \citep{Peyre2016}, in which each iteration involves a projection into the transportation polytope, efficiently computed with the Sinkhorn-Knopp algorithm \citep{Cuturi2013}, a matrix-scaling procedure that alternatingly updates marginal scaling variables. 
In the limit of vanishing regularization ($\epsilon\rightarrow 0$) these scaling factors diverge, resulting in numerical instabilities.

To improve the numerical stability, we compute $\text{GW}_{\epsilon}$ using a stabilized version of the Sinkhorn algorithm \citep{Schmitzer2016}. This significantly increases the robustness of the Gromov-Wasserstein computation. Performing Sinkhorn updates in the log-domain further increases the stability of the algorithm, by avoiding numerical overflow while preserving its efficient matrix multiplication structure.

Normalizing the intra-space distances of the generated and the data samples, respectively, further improves the numerical stability of the Gromov-Wasserstein computation. However, to preserve information on the scale of the samples, we use normalized distances for the Sinkhorn iterates, while the final loss is calculated using the original distances.

\section{Empirical Results}
In this section, we empirically demonstrate the effectiveness of the \textsc{Gw Gan} formulation and regularization, and illustrate its versatility by tackling various novel settings for generative modeling, including learning distributions across different dimensionalities, data types and styles.

\subsection{Learning across Identical Spaces}
As a sanity check, we first consider the special case where the two distributions are defined on identical spaces (i.e., the usual GAN setting). Specifically, we test the model's ability to recover 2D mixtures of Gaussians, a common proof of concept task for mode recovery \citep{Che2016, Metz2017, LiMelis2018}. 
For the experiments on synthetic datasets, generator and adversary architectures are multilayer perceptrons (MLPs) with ReLU activation functions.
Figure~\ref{fig:distributions}a shows that the \textsc{Gw Gan} reproduces a mixture of Gaussians with learned adversary $f_\omega$ that stabilizes the learning. 
We observe that $\ell_1$-regularization indeed helps position the learned distributions around the origin. Comparative results with and without $\ell_1$-regularization are shown in Appendix~\ref{sec:comp_l1}.
As opposed to the \textsc{Ot Gan} proposed by \citet{Salimans2016}, our model robustly learns Gaussian mixtures with differing number of modes and arrangements (see Appendix~\ref{sec:comp_sal}). The Appendix shows several training runs. While the generated distributions vary in orientation in the Euclidean plane, the cluster structure is clearly preserved.

To illustrate the ability of the \textsc{Gw Gan} to generate images, we train the model on MNIST \citep{LeCun1998}, fashion-MNIST \citep{Xiao2017} and gray-scale CIFAR-10 \citep{Krizhevsky2014}.
Both generator and adversary follow the deep convolutional architecture introduced by \citet{Chen2016}, whereby the adversary $f_\omega$ maps into $\mathbb{R}^s$ rather than applying a final \texttt{tanh}.
To stabilize the initial training phase, the weights of the adversary network were initialized with random orthogonal matrices as proposed by \citet{Saxe2013}. 
We train the model using Adam with a learning rate of $2 \times 10^{-4}$, $\beta_1 = 0.5$, $\beta_2 = 0.99$ \citep{Kinga2015}.
Figure~\ref{fig:distributions}b, c and d display generated images throughout the training process.  
The adversary was constrained to approximate an orthogonal operator. The results highlight the effectiveness of the orthogonal Procrustes regularization, which allows successful learning of complex distributions using different network architectures.
Additional experiments on the influence of adversary $f_\omega$ are provided in Appendix~\ref{sec:comp_adv}.

Having validated the overall soundness of the \textsc{Gw Gan} on traditional settings, we now demonstrate its usefulness in tasks that go beyond the scope of traditional generative adversarial models, namely, learning across spaces that are not directly comparable.

\begin{figure*}[t]
  \centering
  \includegraphics[width=.75\textwidth]{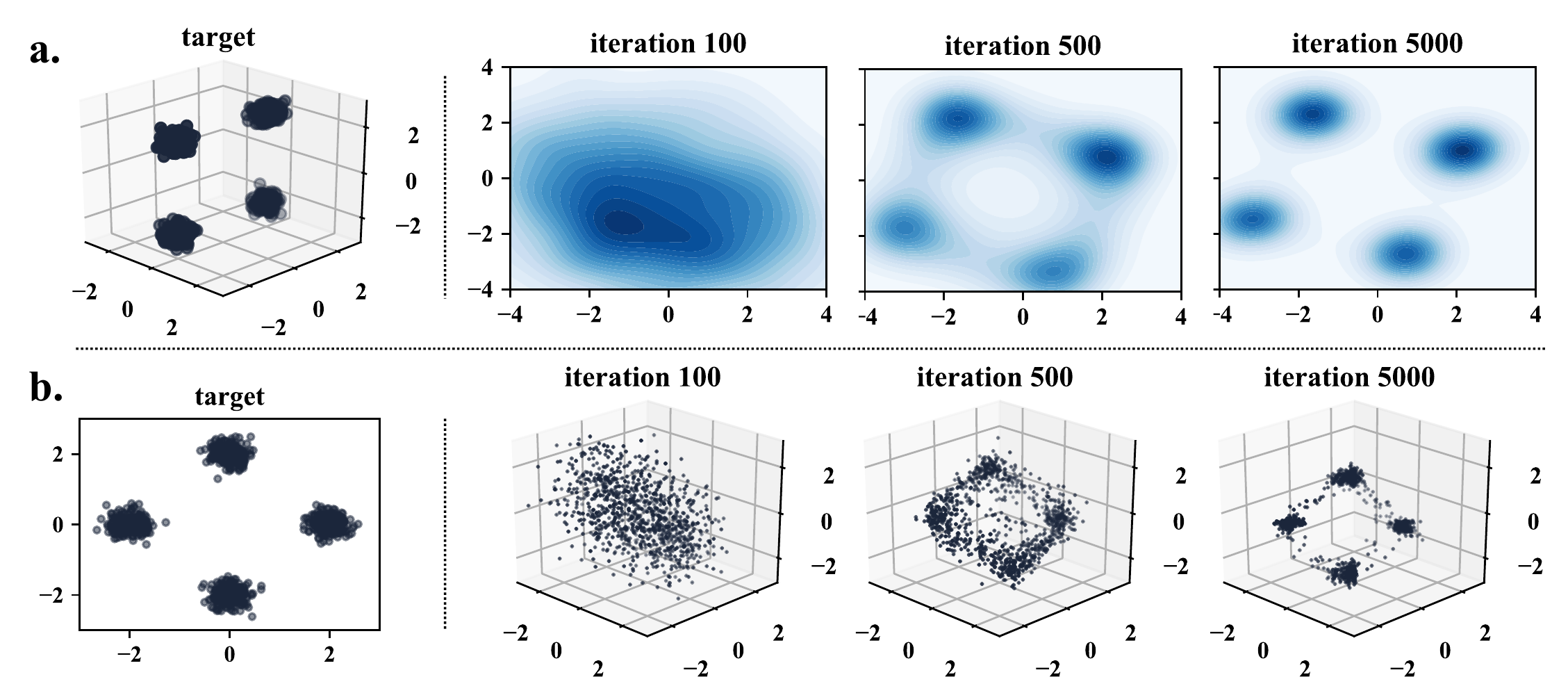}
  \caption{The \textsc{Gw Gan} can be applied to generate samples of \textbf{a.} reduced and \textbf{b.} increased dimensionality compared to the target distribution. All plots show 1000 samples.}
  \label{fig:dimensions}
\end{figure*}
\begin{figure*}[t]
  \centering
  \includegraphics[width=.9\textwidth]{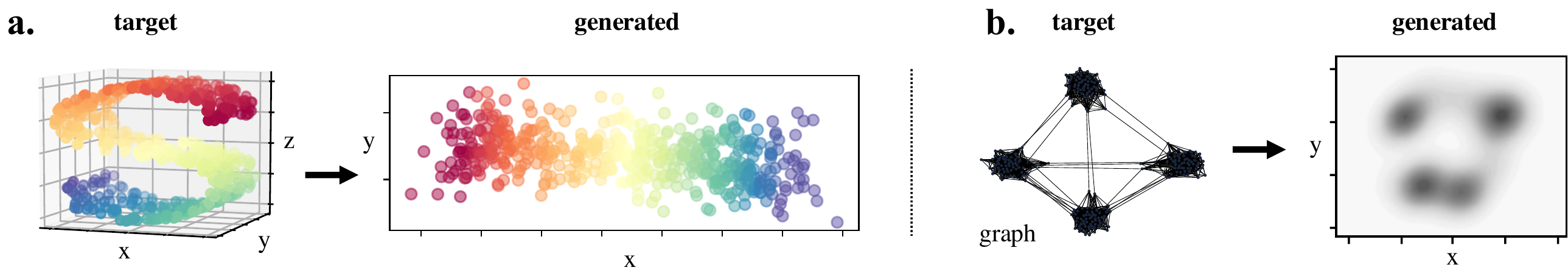}
  \caption{By learning from intra-space distances, the \textsc{Gw Gan} learns the manifold structure of the data. \textbf{a.} The model can be applied to dimensionality reduction tasks and reproduce a three-dimensional S-curve in two-dimensions. Intra-space distances of the data samples are Floyd-Warshall shortest paths of the corresponding k-nearest neighbor graph. \textbf{b}. Similarly, it can map a graph into $\mathbb{R}^2$. The plots display 500 samples.} 
  \label{fig:manifold}
\end{figure*}

\subsection{Learning across Dimensionalities}
Arguably, the simplest instance of \emph{incomparable} spaces are Euclidean spaces of different dimensionality. In this section, we investigate whether the Gromov-Wasserstein GAN can learn to generate a distribution defined on a space of different dimensionality than that of the reference. We consider both directions: learning to a smaller and higher dimensional space.
In this experimental setup, we compute intra-space distances using the Euclidean distance without a parametrized adversary. The generator network follows an MLP architecture with ReLU activation functions. The training task consists of translating between a mixture of Gaussian distributions in two and three dimensions.
The results, shown in Figure~\ref{fig:dimensions}, demonstrate that our model successfully recovers the global structure and relative distances of the modes of the reference distribution, despite the different dimensionality. 

\subsection{Learning across Data Modalities and Manifolds}
Next, we consider distributions with more complex structure, and test whether our model is able to \emph{recover} manifold structure on the generated distribution. Using the popular three-dimensional S-shaped dataset as example, we define distances between the samples via shortest paths on their k-nearest neighbor graph, computed using the Floyd-Warshall algorithm \citep{Floyd1962}. For the generated distribution we use a space of the same intrinsic dimensionality (two) as the reference manifold. The results in Figure~\ref{fig:manifold}a show that the generated distribution learnt with \textsc{Gw Gan} successfully recovers the manifold structure of the data.

Taking the notion of incomparability further, we next consider a setting when the reference distribution is accessible only through relational information, i.e., a weighted graph without absolute representations of the samples. While conceptually very different from previous scenarios, applying our model to this setting is just as simple as previous scenarios.
Once a notion of distance is defined over the reference graph, our model learns the distribution based on pairwise relations as before. 
Given merely a graph, we use pairwise shortest paths as the intra-space distance metric, and use the 2D Euclidean space for the generated distribution. Figure~\ref{fig:manifold}b shows that \textsc{Gw Gan} is able to successfully learn a distribution that approximately recovers the neighborhood structure of the reference graph. 

\subsection{Shaping Learned Distributions}\label{sec:style_bold}
The Gromov-Wasserstein GAN enjoys remarkable flexibility, allowing us to actively influence stylistic characteristics of the generated distribution.

\begin{figure}[t]
  \centering
  \includegraphics[width=.95\columnwidth, trim = {0 1.15cm 0 0}, clip]{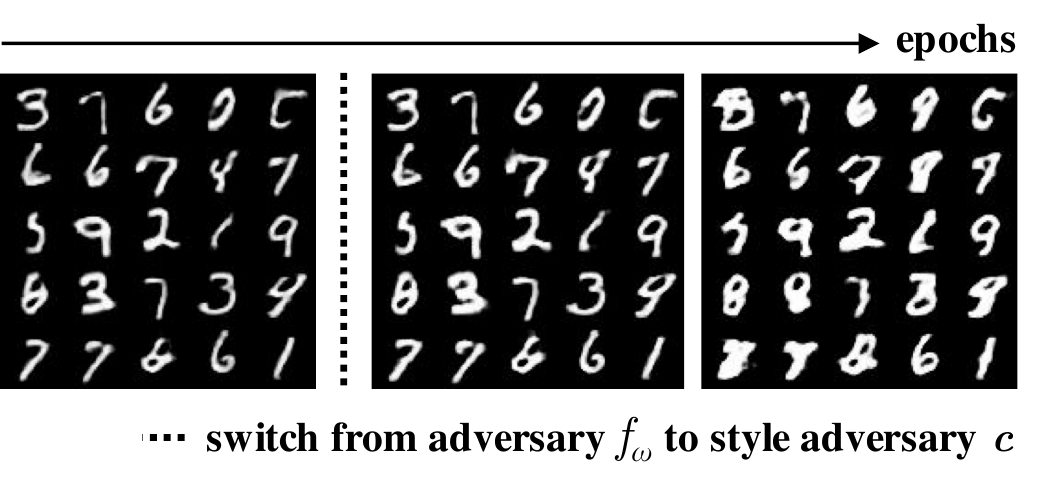}
  \caption{\textbf{Cross-Style Generation}. By decoupling topological information from superficial characteristics, our method allows for stylistic aspects to be enforced (boldness in this case, enforced with a style adversary) upon the generated distribution \emph{while} preserving the principal characteristics of the reference distribution (MNIST).}
  \label{fig:style_transfer}
\end{figure}

While structure and content of the distribution are learned via the adversary $f_\omega$, stylistic features can be introduced via a style adversary as outlined in Section~\ref{sub:constr_gen}.
As a proof of concept of this modular framework, we learn MNIST digits and enforce their font style to be bold via additional design constraints. The style adversary is parametrized by a binary classifier trained on handwritten letters of the EMNIST dataset \citep{Cohen2017} which were assigned thin and bold class labels $l \in \left \{ 0, 1 \right \}$. The training objective of the generator $g_\theta$ is augmented with the classification result of the trained binary classifier (Eq.~\eqref{eq:loss_style}). Further details are provided in Appendix~\ref{sec:train_sa}.
After the generator has satisfactorily learnt the data distribution based on training with loss $\overline{\text{GW}}_\epsilon$, the style adversary $c$ is activated. Figure~\ref{fig:style_transfer} shows that the style adversary affects the generator to increase the thickness of the MNIST digits, while the structural content learned in the first stage is retained.

\section{Related Work}
Generative adversarial models have been extensively studied and applied in various fields including image synthesis \citep{Brock2019}, semantic image editing \citep{Wang2018}, style transfer \citep{Zhu2017}, and semi-supervised learning \citep{Kingma2014}. 
As the literature is extensive, we provide a brief overview on GANs 
and focus on selected approaches targeting tasks in cross-domain learning.

\paragraph{Generative Adversarial Networks}
\citet{Goodfellow2014} proposed generative adversarial networks (GANs) as a zero-sum game between a generator and a discriminator, which learns to distinguish between generated and data samples.
Despite their success and improvements in optimization, the training of GANs is difficult and unstable \citep{Salimans2016, Arjovsky2017a}. 
To remedy these issues, various extensions of this framework have been proposed, most of which seek to replace the game objective with more stable or general losses. These include using Maximum Mean Discrepancy (MMD) \citep{Dziugaite2015, Li2017, Binkowski2018}, other IPMs \citep{Mroueh2017, Mroueh2018}, or Optimal Transport distances \citep{Arjovsky2017, Salimans2018, Genevay2018}. Due to their relevance, we discuss the latter in detail below. A crucial characteristic that distinguishes our approach from other generative models is its ability to learn across different domains and modalities.

\vspace{-5pt}
\paragraph{GANs and Optimal Transport (OT)}
To compare probability distributions supported on low dimensional manifolds in high dimensional spaces, 
recent GAN variants integrate OT metrics in their training objective \citep{Arjovsky2017, Salimans2018, Genevay2018, Gulrajani2017}.
Since OT metrics are computationally expensive, \citet{Arjovsky2017} use the dual formulation of the 1-Wasserstein distance. Other approaches approximate the primal via entropically smoothed generalizations of the Wasserstein distance \citep{Salimans2018, Genevay2018}. 
Our work departs from these methods in that it relies on a much more general instance of Optimal Transport (the Gromov-Wasserstein distance) as a loss function, which allows us to compare distributions even if cross-domain pairwise distances are not available. 
\vspace{-5pt}
\paragraph{GANs for Cross-Domain Learning}
GANs have been successfully applied to style transfer between images \citep{Isola2017, Karacan2016, Zhu2017}, text-to-image synthesis \citep{Reed2016, Han2017}, visual manipulation \citep{Zhu2016, Engel2018} or font style transfer \citep{Azadi2018}. However, to achieve this, these methods depend on conditional variables, training sets of aligned data pairs or cycle consistency constraints.
\citet{Kim2017} utilize two different, coupled GANs to discover cross-domain relations given unpaired data. However, the method's applicability is limited as all images in one domain need to be representable by images in the other domain.
\vspace{-5pt}
\paragraph{Gromov-Wasserstein Learning}
Since its introduction by \citet{Memoli2011}, the Gromov-Wasserstein discrepancy has found applications in many learning problems that rely on a coupling between different metric spaces. Being an effective method to solve matching problems, it has been used in shape and object matching \citep{Memoli2009, Memoli2011, Solomon2016, Ezuz2017}, for aligning word embedding spaces \citep{Melis2018} and for matching weighted directed networks \citep{Chowdhury2018}. Other recent applications of the $\text{GW}$ distance include the computation of barycenters of a set of distance or kernel matrices \citep{Peyre2016} and heterogeneous domain adaptation where source and target samples are  represented in different feature spaces \citep{Yan2018}. While relying on a shared tool ---the $\text{GW}$ discrepancy--- this paper leverages it in a very different framework, generative modeling, where questions of efficiency, degrees of freedom, minimax objectives and end-to-end learning pose various challenges that need to be addressed to successfully use this tool.
\vspace{-7pt}
\section{Conclusion}
In this paper, we presented a new generative model that can learn a distribution in a space that is different from, and even \emph{incomparable} to, that of the reference distribution. Our model accomplishes this by relying on relational ---rather than absolute--- comparisons of samples via the Gromov-Wasserstein distance. Such disentanglement of data and generator spaces opens up a wide array of novel possibilities for generative modeling, as portrayed by our experiments on learning across different dimensional representations and learning across modalities (weighted graph to Euclidean representations). Validated here through simple experiments on digit thickness control, the use of crafted regularization losses on the generator to impose certain stylistic characteristics makes for an exciting avenue of future work.

\section*{Acknowledgements} This research was supported in part by NSF CAREER Award 1553284 and The Defense Advanced Research Projects Agency (grant number YFA17 N66001-17-1-4039). The views, opinions, and/or findings contained in this article are those of the author and should not be interpreted as representing the official views or policies, either expressed or implied, of the Defense Advanced Research Projects Agency or the Department of Defense. Charlotte Bunne was supported by the Zeno Karl Schindler Foundation. We thank Suvrit Sra for a question that initiated this research, and MIT Supercloud and the Lincoln Laboratory Supercomputing Center for providing computational resources.

\bibliography{references}
\bibliographystyle{icml2019}

\newpage
\onecolumn
\section*{Appendix}

\appendix
\section{Training of the \textsc{Gw Gan} with Fixed Adversary}\label{sec:fixed_ad}
\begin{figure}[h!]
	\centering
	\includegraphics[width=.9\textwidth]{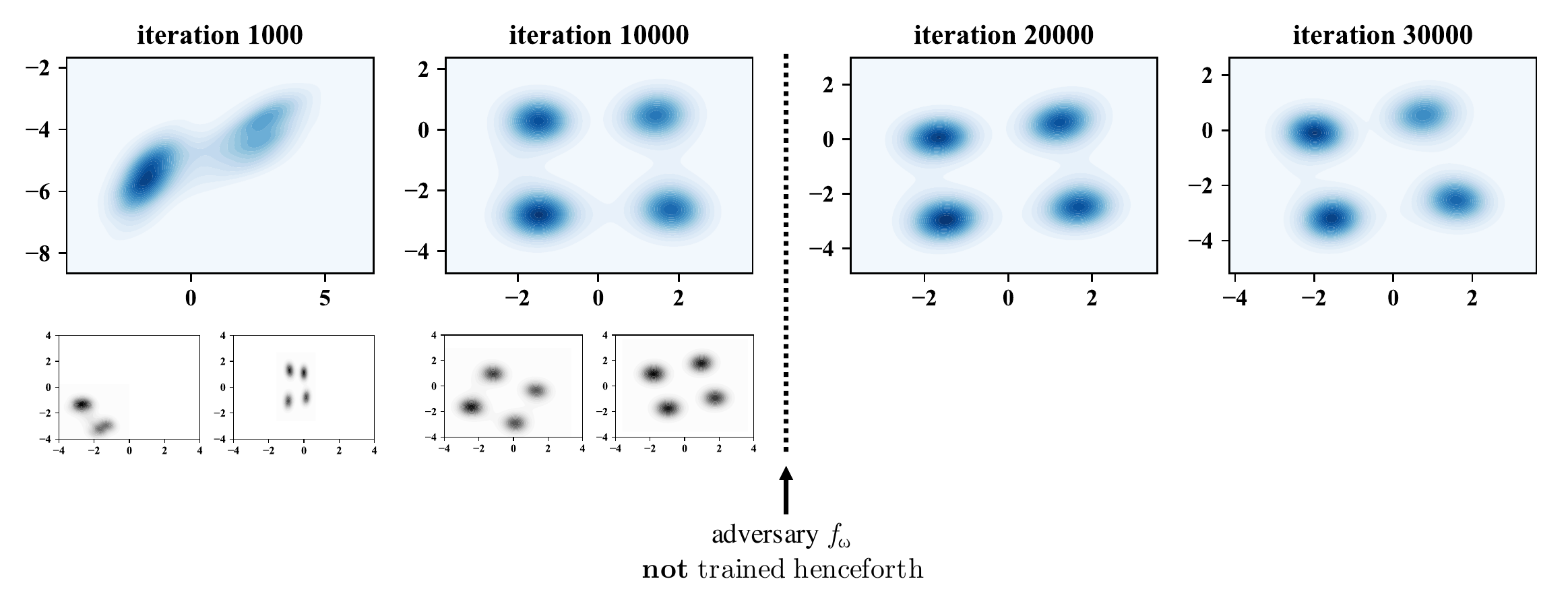}
	\caption{Results demonstrating the consistency of the $\text{GW}$ training objective when fixing adversary $f_\omega$. Even after holding the adversary $f_\omega$ fixed and stopping its training after $10000$ iterations, the generator does not diverge and remains consistent. All plots display 1000 samples.}
	\label{}
\end{figure}
\vspace{7em}
\section{Influence of Generator Constraints on the \textsc{Gw Gan}}\label{sec:comp_l1}
\begin{figure}[h]
	\centering
	\includegraphics[width=.92\textwidth]{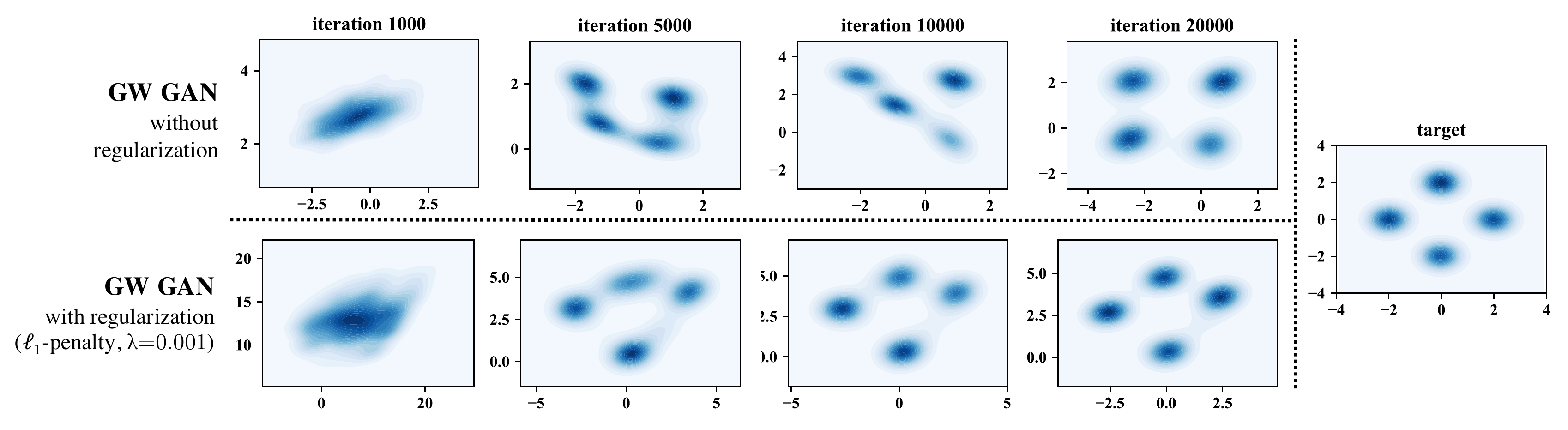}
	\caption{The \textsc{Gw Gan} recovers relational and geometric properties of the reference distribution. Global aspects can be determined via constraints of the generator $g_\theta$. When learning a mixture of four Gaussians we can enforce centering around the origin by penalizing the $\ell_1$-norm of the generated samples. The results show training the \textsc{Gw Gan} with and without a $\ell_1$-penalty. While the \textsc{Gw Gan} recovers all modes in both cases, learning with $\ell_1$-regularization centers the resulting distribution around the origin and determines its orientation in the Euclidean plane. All plots display 1000 samples.}
	\label{}
\end{figure}

\section{Influence of the Adversary}\label{sec:comp_adv}
\begin{figure}[H]
	\centering
	\includegraphics[width=.7\textwidth]{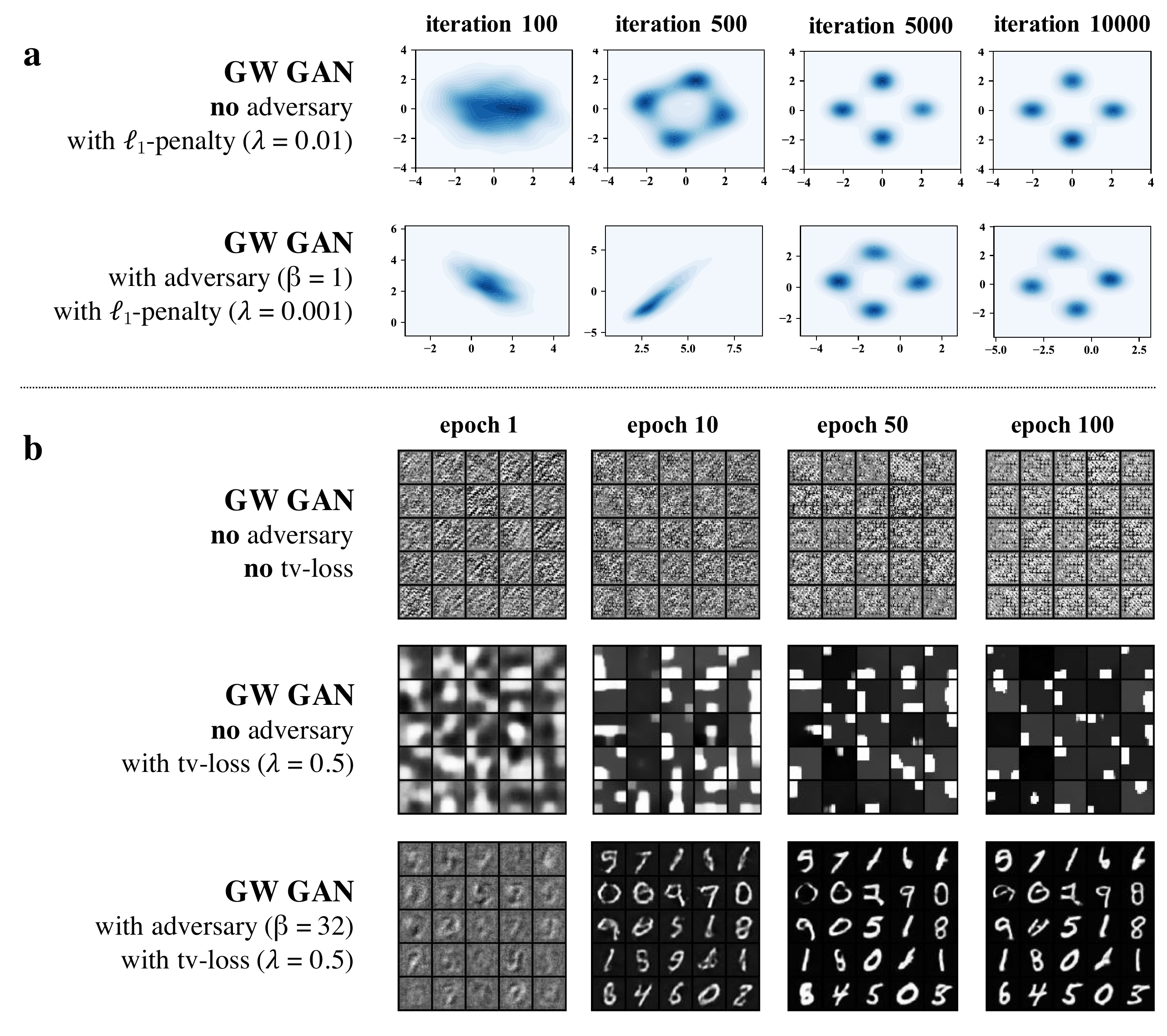}
	\caption{Learning the intra-space metrics adversarially is crucial for high dimensional applications. \textbf{a.} For simple applications such as generating 2D Gaussian distributions, the \textsc{Gw Gan} performs well irrespective of the use of an adversary. \textbf{b.} However, for higher dimensional inputs, such as generating MNIST digits, the use of the adversary is crucial. Without the adversary, the \textsc{Gw Gan} is not able to recover the underlying target distributions, while the total variation regularization (tv-loss) effects a clustering of local intensities.}
	\label{}
\end{figure}

\newpage
\section{Comparison of the Effectiveness of Orthogonal Regularization Approaches}\label{sec:comp_orth}
\begin{figure}[ht!]
	\centering
	\includegraphics[width=.92\textwidth]{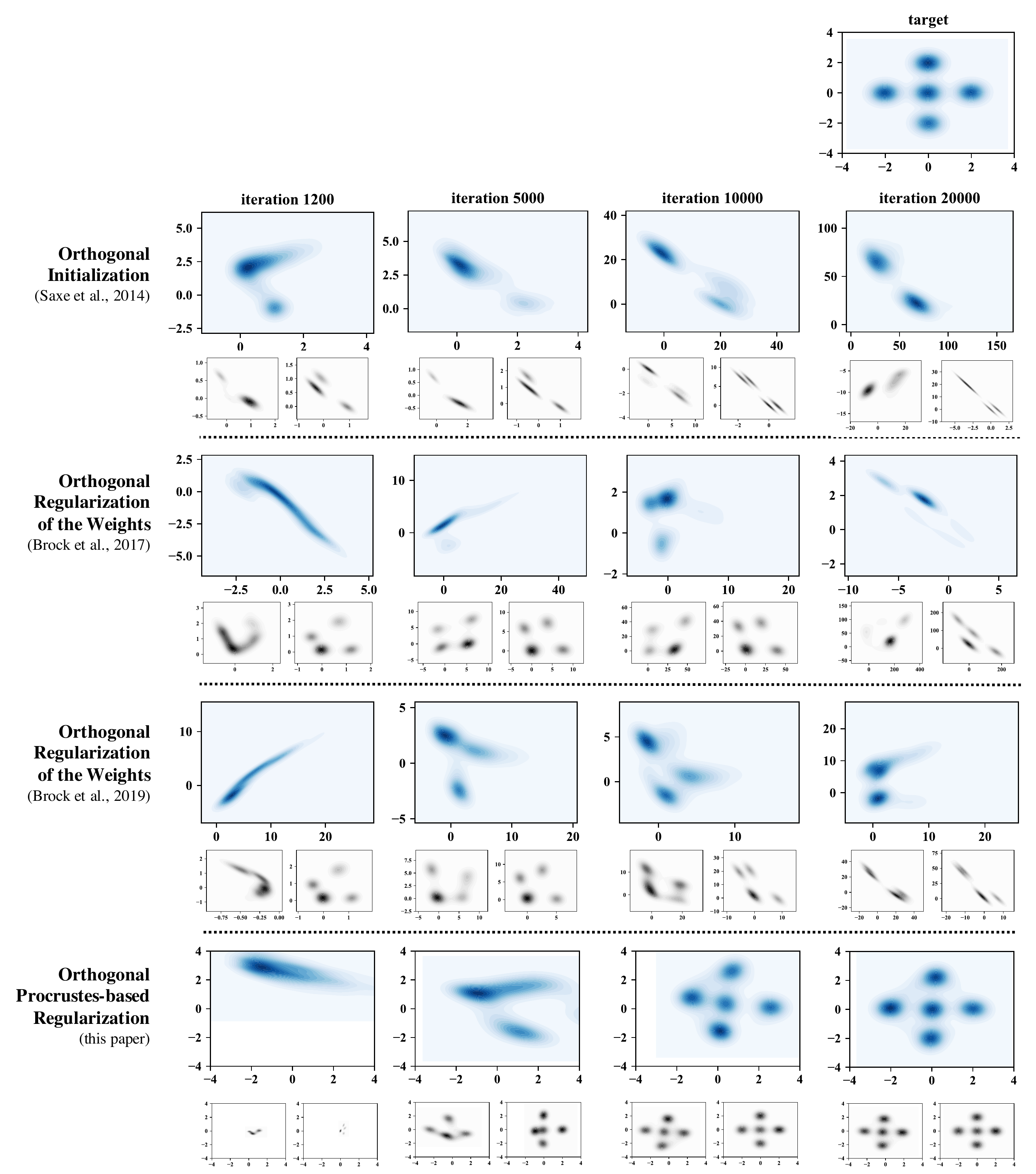}
	\caption{To avoid arbitrary distortion of the space by the adversary, we propose to regularize $f_\omega$ by (approximately) enforcing it to define a unitary transformation. We compare different methods of orthogonal regularization of neural networks; including random initialization of the network's weights at the beginning of the training \citep{Saxe2013}, and layerwise orthogonality constraints which penalize deviations of the weights from orthogonality ($R_\beta(W_k) := \beta \|W_k^\top W_k - I\|_\textrm{F}^2$) \citep{Brock2017}. \citet{Brock2019} remove the diagonal terms from the regularization, which enforces the weights to be orthogonal but does not constrain their norms ($R_\beta(W_k) := \beta \|W_k^\top W_k \odot (1 - I) \|_\textrm{F}^2$). The approaches of \citet{Saxe2013, Brock2017, Brock2019} are not able to tightly constrain $f_\omega$. As a result, the adversary is able to stretch the space and thus maximize the Gromov-Wasserstein distance. Only the orthogonal Procrustes-based orthogonality approach introduced in this paper is able to effectively regularize adversary $f_\omega$ preventing arbitrary distortions of the intra-space distances in the feature space. All plots display 1000 samples.}
	\label{}
\end{figure}

\section{Comparison of the \textsc{Gw Gan} with \citet{Salimans2018}}\label{sec:comp_sal}
\begin{figure}[ht!]
	\centering
	\includegraphics[width=.9\textwidth]{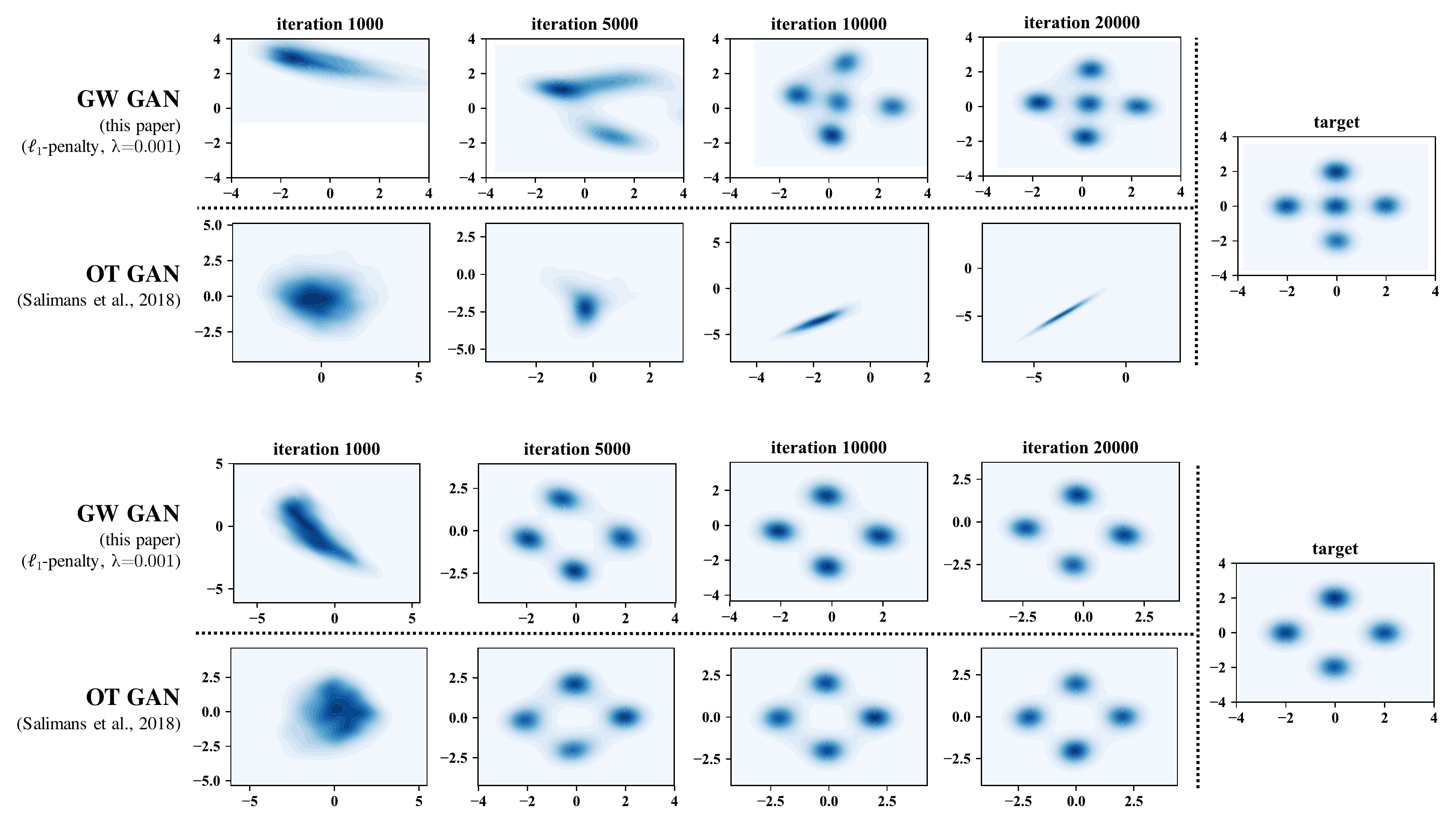}
	\caption{Comparison of the performance of the \textsc{Gw Gan} and an OT based generative model proposed by \citet{Salimans2018} (\textsc{Ot Gan}). The \textsc{Gw Gan} learns mixture of Gaussians with differing number of modes and arrangements. The approach of \citet{Salimans2018} is not able to recover a mixture of five Gaussians with a centering distribution. In the case of a mixture of four Gaussian distributions, both models are able to recover the reference distribution in a similar number of iterations. All plots display 1000 samples.}
	\label{}
\end{figure}

\section{Comparison of Training Times}\label{sec:comp_times}
\begin{table}[ht!]
\centering
\begin{tabular}{lc}
	\textbf{Model} & \textbf{Average Training Time} (Seconds per Epoch) \\
	  \hline
	\textsc{Wasserstein Gan} with Gradient Penalty \cite{Gulrajani2017} & $17.57 \pm 2.07$ \\
  	\textsc{Sinkhorn Gan} \cite{Genevay2018} (default configuration, $\epsilon = 0.1$) & $ 145.52 \pm 1.90$ \\
  	\textsc{Sinkhorn Gan} \cite{Genevay2018} (default configuration, $\epsilon = 0.005$) & $ 153.86 \pm 1.64$ \\
  	\textsc{Gw Gan} (this paper, $\epsilon = 0.005$) & $156.62 \pm 1.06$ 
\end{tabular}
\label{}
\caption{Training time comparisons of PyTorch implementations of different GAN architectures. The generative models were trained on generating MNIST digits and their average training time per epoch was recorded. All experiments were performed on a single GPU for consistency.}
\end{table}

\newpage
\section{Training Details of the Style Adversary}\label{sec:train_sa}
\begin{figure}[H]
	\centering
	\includegraphics[width=0.8\textwidth]{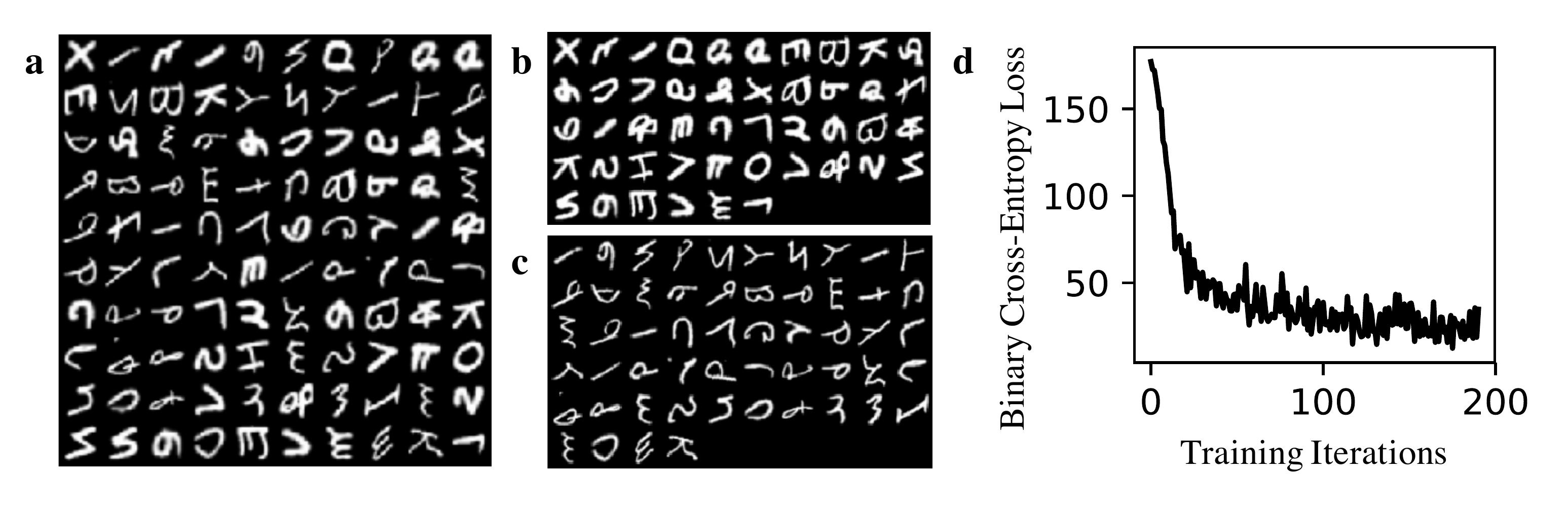}
	\caption{Training of a binary classifier to discriminate between bold and thin letters. \textbf{a.} Training set of the EMNIST dataset including bold and thin letters.  Output of the trained network of letters labelled as \textbf{b.} bold and \textbf{c.} thin. \textbf{d.} Loss curve corresponding to the training of the binary classifier.}
	\label{fig:binary_classifier}
\end{figure}

We introduce a novel framework which allows a modular application of style transfer tasks by integrating a \emph{style adversary} into the architecture of the Gromov-Wasserstein GAN.
In order to demonstrate the practicability of this modular framework, we learn MNIST digits and enforce their font style to be bold via additional design constraints. 
The style adversary is parametrized by a binary classifier trained on handwritten letters of the EMNIST dataset \citep{Cohen2017} which were assigned bold and thin class labels based on the letterwise $\ell_1$-norm of each image.
As the style adversary is trained based on a different dataset, it is independent of the original learning task.
The binary classifier is parametrized by a convolutional neural network and trained by computing a binary cross-entropy loss. The dataset, classification results of bold and thin letters as well as the loss curve of training the binary classifier are shown in figure~\ref{fig:binary_classifier}.%

\end{document}